\documentclass[a4paper]{article}
\usepackage{multirow}
\usepackage{INTERSPEECH2022}
\usepackage{url}

\title{ASR Error Detection via Audio-Transcript entailment}
\name{Nimshi Venkat Meripo, Sandeep Konam}
\address{
  Abridge AI Inc.}
\email{venkatm@abridge.com, san@abridge.com}

\begin{document}

\maketitle
\begin{abstract}
Despite improved performances of the latest Automatic Speech Recognition (ASR) systems, transcription errors are still unavoidable. These errors can have a considerable impact in critical domains such as healthcare, when used to help with clinical documentation. Therefore, detecting ASR errors is a critical first step in preventing further error propagation to downstream applications. To this end, we propose a novel end-to-end approach for ASR error detection using
audio-transcript entailment. To the best of our knowledge, we are the first to frame this problem as an end-to-end entailment task between the audio segment and its corresponding transcript segment. Our intuition is that there should be a bidirectional entailment between audio and transcript when there is no recognition error and vice versa. The proposed model utilizes an acoustic encoder and a linguistic encoder to model the speech and transcript respectively. The encoded representations of both modalities are fused to predict the entailment. Since doctor-patient conversations are used in our experiments, a particular emphasis is
placed on medical terms. Our proposed model achieves classification error rates (CER) of 26.2\% on all transcription errors and 23\% on medical errors specifically, leading to improvements upon a strong baseline by 12\% and 15.4\%, respectively.

\end{abstract}
\noindent\textbf{Index Terms}: error-detection, speech recognition, medical conversations, bert, wav2vec2.0, hubert

\section{Introduction}
%% Motivating the problem paragraph
Electronic Health Record (EHR) documentation results in a high administrative burden for physicians and is a significant cause of physician burnout ~\cite{gesner2019burden}. Automatic Speech Recognition (ASR)  is a promising technology to help physicians with documentation. Various studies have shown that speech recognition for medical dictations increases speed and allows for capturing more comprehensive documentation ~\cite{koivikko2008improvement, vogel2015analysis}. Physicians have expressed increased satisfaction using speech recognition for documenting dictations ~\cite{goss2019clinician}. Beyond dictation solutions, there has also been an uptick in adopting manual scribing solutions to reduce the documentation burden so that physicians can focus more on patient care. More recently, there has been a growing interest in leveraging the ASR transcripts and automating parts of the scribing workflows using spoken conversation understanding technologies such as utterance classification~\cite{schloss2020towards}, information extraction ~\cite{med1,med2,appt}, and conversation summarization ~\cite{summ}. Since speech recognition is upstream of all these dictation and scribing-centered documentation workflows, words mistranscribed by the ASR system ~\cite{goss2016incidence} can be problematic, leading to users performing time-consuming and laborious corrections ~\cite{kumah2018electronic}. Leaving a mistranscribed word uncorrected may result in unclear documentation, embarrassing errors, and risk to patient safety ~\cite{kumah2018electronic}. Therefore, it is becoming increasingly relevant to detect the transcription errors beforehand to prevent them from going into the downstream applications and workflows.

\begin{table}[t]
  \caption{Mistranscriptions of medical terms}
  \label{tab:mere}
  \centering
  \begin{tabular}{ll}
    \toprule
    \textbf{Source}      & \textbf{Transcript}                \\
    \midrule
    Reference  & Keep her on the symbicort for a while. \\
    Hypothesis  & Keep her on the civil court for a while.  \\
    \midrule
    Reference   & Last time I checked did an echocardiogram. \\
    Hypothesis  & Last time I checked identical car drive.  \\
    \midrule
    Reference   & It's September now it's flu season. \\
    Hypothesis  & It's September now is who season. \\
    \bottomrule
  \end{tabular}
\end{table}

\begin{table}[t]
  \caption{Mistranscriptions of non-medical terms}
  \label{tab:ere}
  \centering
  \begin{tabular}{ll}
    \toprule
    \textbf{Source}      & \textbf{Transcript}                \\
    \midrule
    Reference  & Good I'd walk in the doctor's office. \\
    Hypothesis  & Could I walk in a box office.  \\
    \midrule
    Reference  & Probably you won't give a timetable. \\
    Hypothesis  & Probably you want to give a timetable.  \\
    \midrule
    Reference  & It will heal you, go back there a year from now. \\
    Hypothesis  & It will heal you, go back the or a year from now.  \\
    \bottomrule
  \end{tabular}
\end{table}

In recent years, cloud-based speech-to-text services from technology companies such as Google and Amazon have become popular, offering flexibility and convenience. However, while these services exhibit impressive performance for general use-cases, they are not optimal for domain-specific vocabulary ~\cite{mani2020asr, mani, khandelwal20_interspeech}. Moreover, these services are provided as black boxes without access to the inner workings of their ASR models, making it impossible to finetune on domain-specific tasks and datasets. Apart from the mismatch in domains, several factors may affect the performance of ASR systems. In particular, the spontaneity of speech and the presence of multiple speakers in medical conversations can be challenging. Furthermore, in an uncontrolled setting such as a hospital, the background noise can also cause recognition errors. In tables \ref{tab:mere} and \ref{tab:ere}, we show domain-specific and generic ASR errors.

In this paper, we propose a novel approach to ASR error detection by utilizing both audio and text modalities. We fuse an acoustic encoder and linguistic encoder into a single model to predict the entailment between the audio and transcript. In the fused model, we freeze the weights of the linguistic encoder, so the acoustic encoder has to learn discriminative features from speech during the training stage to detect ASR errors. Instead of relying on features from the ASR system, our approach encodes the raw speech audio and the hypothesized transcript for error detection. Since our approach does not depend on the inner workings of the ASR system, it can be applied to the outputs of any ASR model. Furthermore, since our approach uses both acoustic and linguistic features, the model can exploit syntactic, semantic, and acoustic cues. In summary, the primary contribution of this paper is a novel end-to-end approach to ASR error detection based on entailment between audio and transcript segments. To help understand the impact on downstream clinical use-cases, we pay special attention to the medical term transcription errors. To understand the robustness of the proposed model to linguistic variations, we also evaluate our model by simulating ASR errors different from those in the original hypothesis.

\section{Related work} 
ASR error detection is a well-established problem and has been studied extensively. The majority of past research can be classified into two major categories. The first set of approaches only uses features generated from the ASR decoder, such as the posterior probability of a word given an acoustic signal. ~\cite{8639602, jiang2005confidence, errattahi2018automatic}. The second set of approaches focuses on linguistic features from the hypothesized transcription in addition to ASR decoder features ~\cite{chen2013asr, pellegrini2010improving}.

Early error detection research focused on the use of confidence measures to understand the reliability of ASR transcriptions ~\cite{jiang2005confidence}. In ~\cite{allauzen2007error}, the authors propose using a logistic regression model with features extracted from confusion networks as a post-processing step for ASR error detection. Later ~\cite{8639602} uses data mining approaches leveraging features from ASR decoder's confidence scores and linguistic parsing.  ~\cite{chen2013asr} ventures beyond the features from the ASR decoder and hypothesized transcript by using additional features from statistical machine translation, posteriors from an entity detector, and a word boundary detector. In ~\cite{sasr}, the authors propose an ASR system-independent error detection and classification model based on a variant RNN-based model. More recently, ~\cite{gekhman2022red} propose an approach to combine ASR confidence scores and transcription hypotheses using Transformer-based models. Most prior work focuses on extensive feature engineering or relies mainly on the hypothesized transcript. In contrast, our approach doesn't require any feature engineering since we encode the raw audio signal and the hypothesized transcript directly. Also, our approach is the first to combine both audio and transcript features in an end-to-end fashion for ASR error detection.

% Recently, ~\cite{ghannay2016acoustic} used acoustic embeddings and evaluated the discriminative ability of these embeddings for ASR error detection
% , while a few ventured into acoustic embeddings.

In the medical domain, there has been an increase in the adoption of ASR transcripts of medical conversations to automate clinical documentation ~\cite{schloss2020towards, krishna2020generating}. While there has been significant research on ASR error detection in the past, ASR error detection on medical conversations is not explored in depth. In this work, we build on ~\cite{mani2020asr} by integrating acoustic features as the source of truth. However, we focus on error detection rather than error correction. Similar to ~\cite{mani2020asr} our current approach does not depend on ASR model features other than the hypothesized transcriptions. Therefore our error detection system can be used regardless of the transcription source. In this work, we develop an end-to-end ASR error detection system using both acoustic and linguistic representations to evaluate the performance in the framework of medical conversations. To the best of our knowledge, we
are the first to frame this problem as an end-to-end entailment task between the audio segment and its corresponding transcript segment.

% focused on medical conversations and attempted to correct ASR errors using sequence-to-sequence-based approaches. Our approach relied solely on linguistic features, and in this paper, we propose

\section{Error Detection Model}
The aim of ASR error detection is to mitigate the impact of mistranscription by flagging the recognition errors automatically. We model the ASR error detection problem as binary entailment prediction task between hypothesized text and speech signal where positive entailment indicates that the speech signal and corresponding hypothesis text imply each other and a negative entailment indicates they contradict each other. Thus the positive entailment implies no error, and the negative entailment implies an error.  In our approach, we use an acoustic encoder and a linguistic encoder to encode the speech and corresponding transcript respectively. Further, we fuse the representations of these two modalities and pass them to an entailment classifier to predict the entailment. Using this model, we aim to automatically detect any systematic speech recognition errors, including medical domain-specific terms.

\subsection{Acoustic Encoders (AE)}
We experiment with Wav2vec2.0 ~\cite{baevski2020wav2vec} and HuBERT ~\cite{ hsu2021hubert} separately as acoustic encoders in our model since they have been well studied and verified. They also have the advantage of operating directly on speech signals without requiring any feature engineering.

\subsubsection{Wav2vec2.0}
Wav2vec2.0 is a framework for self-supervised learning of speech representations. Taking inspiration from masked language modeling, wav2vec2.0 encodes audio speech using convolutional layers into a latent space representation $Z$ and masks the spans in latent space. % $Z$ represents the raw audio sampled at 16k with a stride of about 20ms and a receptive field of 25ms. 
The latent representations $Z$ are then forwarded to a transformer network to obtain a contextual representation $C$. The model is then pretrained via solving a contrastive task. Wav2vec2.0 outperforms previous semi-supervised methods simply by
fine-tuning on transcribed speech with connectionist temporal classification criterion ~\cite{baevski2020wav2vec}. In this work, wav2vec2.0 is used to convert audio speech signals to an acoustic representation. We particularly use the contextualized representations $C$ as our acoustic representation.

\subsubsection{HuBERT}
HuBERT is a self-supervised speech representation learning approach and follows an architecture similar to Wav2vec2.0. It has state of the art performance either matching or improving upon Wav2vec2.0. The speech representations are learned by pre-training with a masked prediction task. HuBERT uses an offline k-means clustering step and learns the speech representations by alternating between clustering and prediction processes by predicting the proper cluster for the masked audio signal. It relies mainly on the consistency of the unsupervised clustering step rather than the intrinsic quality of the allocated cluster labels ~\cite{ hsu2021hubert}.
% HuBERT can also be used as an encoder to convert raw speech signals to acoustic representations. 

\subsection{Linguistic Encoder (LE)}
We choose BERT ~\cite{devlin2018bert} as the linguistic encoder in our model. BERT is the language model trained using masked language modeling objective. During pre-training BERT utilizes both the left and right context and tries to predict the masked words. BERT is trained on large amounts of text data and pre-trained BERT has been shown to achieve impressive performance even in cases of limited data for downstream tasks. Previous probing analyses have also shown that pretrained BERT captures syntactic dependency structures and coreferences with remarkable accuracy ~\cite{penha2020does, wu2020perturbed}. Thus, the priors on BERT can provide a robust semantic and syntactic context in encoding the ASR hypothesis.
% The BERT model is based on the transformer architecture and is composed of fully connected and self-attention layers. An important advantage of the BERT model is that it operates on subword units, so it can also handle out of vocabulary words.  

\subsection{Entailment Classifier}
Since the speech signal and transcript can be of varying lengths, the representations obtained from the acoustic encoder and linguistic encoder are averaged to a fixed dimensional representation. We also experimented with the self-attention mechanism to obtain fixed-size representations, but found these did not yield better results than simple averaging.

$$a_{i} = W_{a} . AE(s_{i})  + b_{a}$$
$$l_{i} = W_{l} . LE(t_{i})  + b_{l}$$
$$c_{i} = concat([a_{i}, l_{i}])$$
$$ e_{i} = sigmoid(W_{e} c_{i} + b_{e}) \ \ \forall i=1...M $$

Because the representations from acoustic and linguistic encoders are from different modalities, we use a projection layer to bring these modalities to the same representation space. Let  $s_{i}$ be the speech segment normalized with zero-mean unit-variance and $t_{i}$ be the hypothesized transcript segment. We obtain the representations $a_{i}$ and $l_{i}$ from the projection layer as shown in the above equations. We then concatenate these projected representations to obtain the context vector $c_{i}$. Finally, the context vector is applied to a fully connected layer followed by a sigmoid layer to predict the entailment.

% We believe utterance level context cues, for this reason we model the errors at utterance level rather than word level. 

\section{Experiments}
\subsection{Dataset}\label{subsec:Dataset}
Using reference human-written and ASR transcripts of the same speech signal, we automatically label the recognition errors and construct a dataset of speech and ASR transcript segments. We use a fully-consented and deidentified dataset containing 3807 doctor-patient conversations with an average conversation length of 9 minutes. %The conversations come from various medical specialties, including cardiology, rheumatology, primary care, and internal medicine. 

The ASR hypothesis is generated using the Google Speech-to-Text API service \footnote{\url{https://cloud.google.com/speech-to-text}}. For comparison, we show the error rates computed on the same dataset of hypothesized transcripts in Table \ref{tbl:wer} from our previous work ~\cite{mani2020asr}. We use the sentence tokenizer from Spacy ~\cite{spacy2} to chunk both the reference and hypothesis transcripts into segments. Google Speech-to-Text API provided word-level timestamps can be used to align reference and hypothesis segments. However, the timestamps in the human-annotated reference transcripts are often off by a few seconds.

\paragraph*{Alignment} Since the timestamps in the reference transcript are not reliable, specific word-level alignment techniques are required to match the output of the ASR system. The alignment is achieved using the Smith-Waterman algorithm, which obtains optimal local alignments between reference and hypothesis transcripts. A lower gap penalty and higher match score is used to favor local alignments when possible. To obtain good alignments, a series of text cleaning steps, casing, stemming, and punctuation stripping are performed before applying the alignment algorithm.
\medbreak
After the alignment process, we perform simple text matching to label ASR errors. Since we focus on downstream clinical documentation applications, we also label ASR errors particularly in medical terms. To annotate medical term-related errors, we identify medical terms in aligned reference and hypothesis transcript segments using the biomedical ontology UMLS ~\cite{schuyler1993umls} and perform string matching.

\paragraph*{UMLS}
    UMLS is a biomedical ontology with comprehensive coverage of medical terms often used in clinical information extraction systems. The medical terms are manually categorized and organized into semantic groups such as body system, clinical drug, etc. For evaluation purposes, we consider five major semantic groups: Chemicals \& Drugs, Disorders, Procedures, Anatomy, and Physiology. To identify medical terms in each transcript segment, we extract n-grams from the transcript and apply a dictionary matching algorithm to match the medical terms in the UMLS database. With the diversity of clinical specialties in the dataset, we obtain segments with various medical terms. 

% \medbreak
In this study, we train our error detection models at the segment level rather than word level for two reasons. First, interpolating word-level timestamps for ASR errors may not be reliable, and second, transcript segments can provide valuable contextual cues regarding recognition error. To allow for a fair comparison between the error detection task and the domain-specific error detection task, we balance the datasets for each type of error and finally obtain nearly 100 hours of speech data with 65k speech segments and corresponding hypothesized transcript segments and labels. The minimum and maximum length of audio segments are chosen to be between 1 and 30 seconds. We set aside 80\% of the data for training purposes, and the remaining 20\% data is split equally for validation and testing. To evaluate the robustness of the model to linguistic variations, we construct a simulated error test set by replacing mistranscribed words in the test set using the ASR error simulation approach proposed in ~\cite{voleti2019investigating}. 

We use base models pre-trained on the librispeech 960-hour dataset for acoustic encoders. For the linguistic encoder, we use the uncased base BERT model. While training, we freeze the weights of the linguistic encoder and latent representation extractor network in the acoustic encoder. The contextual Transformer network in the acoustic encoder is fine-tuned while training the entailment classifier. The intuition is that the BERT model provides strong syntactic and semantic context. At the same time, the acoustic encoder has to learn representations specific to the error detection task to entail the linguistic representation. 

We construct a baseline model by replacing the acoustic encoder in our proposed model with ASR confidence scores as features (CS + BERT). 

We use a learning rate of 1e-3 to train all models.
% To compare against, we construct two baselines. The first one using a threshold on ASR confidence scores \footnote{Threshold is chosen based on optimal F1 score on development set} and the second one is replacing acoustic encoder in our proposed model with confidence score features.

\begin{table}[t]
\caption{Word error rate and medical word error rates on hypothesized transcripts}% (Medical WER).}
\centering
\begin{tabular}{llccc}
\toprule
\multirow{2}{*}{\textbf{WER}}  &  \multirow{2}{*}{\textbf{BLEU}} & \multicolumn{3}{c}{\textbf{Medical WER}}      \\
\cmidrule{3-5}
&   &  P  & R & F1  \\ 
\midrule
41.0  & 52.1 & 90.1  & 58.4  & 67.3   \\

\bottomrule
\end{tabular}
\label{tbl:wer}
\end{table}

% \clearpage

\begin{table}[hbt!]
\caption{Evaluation of error detection entailment task using confidence scores (CS) and acoustic features}% (Medical WER).}
\centering
\begin{tabular}{lcccc}
\toprule
\textbf{Model} & \textbf{Precision} & \textbf{Recall} & \textbf{F1} & \textbf{CER} \\
\midrule
\multicolumn{5}{c}{Medical Errors} \\
\midrule
% CS & 54.13 &  74.37 & 62.66   &  47.6 \\
CS + BERT & 72.14 &  80.54 & 76.11   &  27.2 \\
wav2vec2.0 + BERT & \textbf{75.57} &  84.18 & 79.63   &  23.3 \\
HuBERT + BERT & 74.00 &  \textbf{88.24} & \textbf{80.46}   &  \textbf{23.0} \\
\midrule
\multicolumn{5}{c}{All Errors} \\
\midrule
% CS & 48.56 & 70.56   &  57.53 & 52.1 \\
CS + BERT & 70.08 &  70.65 &  70.36   &  29.8 \\
wav2vec2.0 + BERT & \textbf{75.51} & 70.44 & 72.89   &  \textbf{26.2} \\
HuBERT + BERT & 74.28 &	\textbf{72.29} &	\textbf{73.27} &  26.4 \\
\bottomrule
\end{tabular}
\label{tbl:res_error_merror}
\end{table}

\begin{table}[hbt!]
\caption{Evaluation of best performing HuBERT + BERT model on simulated ASR errors}% (Medical WER).}
\centering
\begin{tabular}{lcccc}
\toprule
\textbf{Error Type} & \textbf{Precision} & \textbf{Recall} & \textbf{F1} & \textbf{CER} \\
\midrule
% \multicolumn{5}{c}{Medical Errors} \\
% \midrule
% CS + BERT & 72.14 &  80.54 & 76.11   &  27.2 \\
% wav2vec2.0 + BERT & 75.57 &  84.18 & 79.63   &  23.3 \\
Medical Errors & 71.99 &  91.86 & 80.72   &  23.5 \\
% \midrule
% \multicolumn{5}{c}{All Errors} \\
% \midrule
% CS + BERT & 72.14 &  80.54 & 76.11   &  27.2 \\
% wav2vec2.0 + BERT & 75.51 &  70.44 & 72.89   &  26.2 \\
All Errors & 77.86 &  81.01 & 79.41   &  21.0 \\

\bottomrule
\end{tabular}
\label{tbl:robust}
\end{table}

\section{Results and Discussion}
For evaluation, we choose classification error rate (CER) along with precision, recall, and F1 metrics. We evaluate our model both qualitatively and quantitatively. To understand the impact on downstream clinical documentation applications, we also focus on major semantic groups of medical terms. In addition, we also evaluate the model for robustness by simulating ASR errors.

\paragraph*{Error type detection} We present our results on both medical and all error detection tasks using different acoustic encoders and the baseline confidence score features (CS + BERT) in Table \ref{tbl:res_error_merror}. When compared with the baseline model using confidence scores, we see an improvement across all metrics showing the difficulty of the task and effectiveness of using linguistic and acoustic representations together.  In both the domain-specific and  general error detection tasks, the models using the HuBERT acoustic encoder have improved recall performance, while those using the Wav2vec2.0 acoustic encoder have slightly better precision. Overall, we observe that the domain-specific medical term classification error rate is lower compared to the generic error detection task. We hypothesize that medical terms have specific linguistic and acoustic patterns and hence are easier to recognize than general errors which are more nuanced and can happen in an unpredictable manner. 

\paragraph*{Semantic Groups and Medical Terms} We measure the medical error detection metrics specifically for the five major semantic groups. To compare errors based on semantic groups, we evaluate our chosen metrics on segments containing medical terms of each corresponding semantic group. In Table \ref{tbl:sem_group}, we notice that the ASR error detection performs best on the Chemicals \& Drugs semantic group. This semantic group is the most frequent error type and mainly captures medications. Apart from having slightly better performance on Chemicals \& Drugs semantic group, the metrics are very consistent across the remaining semantic groups. In Table \ref{tbl:qualitative}, we specifically focus on the top five most frequent and top five least frequent medical terms. For the least frequent medical terms, we randomly sample five medical terms at a cut off frequency of ten. We notice a relatively higher performance on frequently occurring medical terms, although the performance can be varied in least frequent terms as the number of samples is low.

\paragraph*{Robustness}
In Table \ref{tbl:robust}, we evaluate the best performing HuBERT + BERT model on the simulated error dataset. On the medical error detection task, the model performs comparably on both original and simulated error datasets. On the other hand, there's a significant improvement of performance on all errors with the simulated dataset suggesting that simulated errors are much easier to detect than the actual ASR hypothesized errors. Overall the metrics suggest that the model can also work well in case of different linguistic variations in ASR errors.

\begin{table}[hbt!]
\caption{Evaluation of medical error detection based on 5 major semantic groups}% (Medical WER).}
\centering
\begin{tabular}{lcccc}
\toprule
\textbf{Semantic Group} & \textbf{Precision} & \textbf{Recall} & \textbf{F1} & \textbf{CER} \\
\midrule
Chemicals \& Drugs & \textbf{92.21} & 90.11 & \textbf{91.15} & \textbf{16.0} \\
Disorders & 86.75 & 91.52 & 89.07 & 19.0  \\
Anatomy & 86.20 & 90.61 & 88.35 & 20.4 \\
Procedures & 85.42 & 91.23 & 88.23 & 20.4 \\
Physiology & 86.19 & \textbf{91.59} & 88.80 & 19.4 \\
\bottomrule
\end{tabular}
\label{tbl:sem_group}
\end{table}

\begin{table}[hbt!]
\caption{Evaluation based on most and least frequent medical term mistranscriptions}% (Medical WER).}
\centering
\begin{tabular}{lcccc}
\toprule
\textbf{Medical term} & \textbf{Precision} & \textbf{Recall} & \textbf{F1} & \textbf{CER} \\
\midrule
\multicolumn{5}{c}{Most frequent} \\
\midrule
heart  & 85.42 & 93.51 & 89.28 & 18.8 \\
blood pressure & 83.09 & 91.47 &  87.08 & 21.4 \\
coumadin &  \textbf{92.59} & 93.47 & 93.02 & 13.0  \\
cholesterol & 86.95 & 97.90 & 92.10 & 14.3 \\
atrial fibrillation &  90.52 & \textbf{98.85} & \textbf{94.50} & \textbf{10.4} \\
\midrule
\multicolumn{5}{c}{Least frequent} \\
\midrule
chest pain & 75.00 & 80.00 & 77.41  & 35.0 \\
stress & 90.90 & 66.66 & 76.92 & 35.3 \\
angiogram & \textbf{93.33} & 73.68 & 82.35 & 30.0 \\
A1c & 88.00 & \textbf{95.65} & \textbf{91.66} & \textbf{15.3} \\
vitamin & 91.30 & 87.50 & 89.36 & 19.2 \\
\bottomrule
\end{tabular}
\label{tbl:qualitative}
\end{table}

\section{Conclusion}
In this work, we introduced a novel approach for automatic ASR error detection in an end-to-end manner. Our approach jointly encodes speech features from audio and linguistic features from transcription hypotheses into a contextualized representation. The entailment classifier detects the ASR errors by modeling features at semantic, syntactic, and acoustic levels. Since our method doesn't rely on any ASR model, it is viable for many domain-specific applications and is easy to implement as a post-processing step. We also evaluated our error detection approach on medical terms and a simulated ASR error dataset for robustness. With good recall on major semantic groups, including disorders and procedures, the proposed model can be a potential candidate for detecting errors. In future work, we would like to investigate pre-training the acoustic feature extractor with domain-specific data. Further, we plan to extend our work for ASR error correction by jointly training a language decoder and evaluate under different acoustic conditions.

\section{Acknowledgements}
We would like to acknowledge Shruti Palaskar for helpful discussions during the preparation of this work and, Elisa Ferracane and Sai Prabhakar Pandi Selvaraj for valuable feedback on the manuscript.

% We also thank the University of Pittsburgh Medical Center (UPMC) and Abridge AI Inc. for providing access to de-identified data.

\newpage
\bibliographystyle{IEEEtran}

\bibliography{template}

% \begin{thebibliography}{9}
% \bibitem[1]{Davis80-COP}
%   S.\ B.\ Davis and P.\ Mermelstein,
%   ``Comparison of parametric representation for monosyllabic word recognition in continuously spoken sentences,''
%   \textit{IEEE Transactions on Acoustics, Speech and Signal Processing}, vol.~28, no.~4, pp.~357--366, 1980.
% \bibitem[2]{Rabiner89-ATO}
%   L.\ R.\ Rabiner,
%   ``A tutorial on hidden Markov models and selected applications in speech recognition,''
%   \textit{Proceedings of the IEEE}, vol.~77, no.~2, pp.~257-286, 1989.
% \bibitem[3]{Hastie09-TEO}
%   T.\ Hastie, R.\ Tibshirani, and J.\ Friedman,
%   \textit{The Elements of Statistical Learning -- Data Mining, Inference, and Prediction}.
%   New York: Springer, 2009.
% \bibitem[4]{YourName17-XXX}
%   F.\ Lastname1, F.\ Lastname2, and F.\ Lastname3,
%   ``Title of your INTERSPEECH 2022 publication,''
%   in \textit{Interspeech 2022 -- 23\textsuperscript{rd} Annual Conference of the International Speech Communication Association, September 18-22, Incheon, Korea, Proceedings, Proceedings}, 2022, pp.~100--104.
% \end{thebibliography}

\end{document}